\documentclass[journal]{IEEEtran}
\usepackage{cite}

\ifCLASSINFOpdf
   \usepackage[pdftex]{graphicx}
   \graphicspath{{../pdf/}{../jpeg/}}
   \DeclareGraphicsExtensions{.pdf,.jpeg,.png}
\else
  \usepackage[dvips]{graphicx}
  \graphicspath{{../eps/}}
  \DeclareGraphicsExtensions{.eps}
\fi
\usepackage{amsmath,amsfonts}
\usepackage{algorithm}
\usepackage{algorithmic}

\usepackage{array,multirow}
\usepackage{booktabs}

\ifCLASSOPTIONcompsoc
  \usepackage[caption=false,font=normalsize,labelfont=sf,textfont=sf]{subfig}
\else
  \usepackage[caption=false,font=footnotesize]{subfig}
\fi
\usepackage{fixltx2e}
\usepackage{url}

\begin{document}
%
\title{Affinity Fusion Graph-based Framework for Natural Image Segmentation}
%
%
%

\author{Yang~Zhang, Moyun~Liu, Jingwu~He, Fei~Pan, and Yanwen~Guo
\thanks{Manuscript received ** **, 2020; revised ** **, 2021. This work was
supported in part by the National Natural Science Foundation of China
(Grant 62032011, 61772257 and 61672279), the Fundamental Research Funds for the Central Universities 020214380058, and the program B for Outstanding PhD candidate of Nanjing University 202001B054. \emph{(Corresponding author: Yanwen~Guo.)}}
\thanks{Y. Zhang, J. He, F. Pan, and Y. Guo are with the National Key Laboratory for Novel Software Technology, Nanjing University, Nanjing 210023, China (e-mail: yzhangcst@smail.nju.edu.cn; hejw005@gmail.com; felix.panf@outlook.com; ywguo@nju.edu.cn).}
\thanks{M. Liu is with the School of Mechanical Science and Engineering, Huazhong University of Science and Technology, Wuhan 430074, China (e-mail: lmomoy8@gmail.com).}
}

%
%

\markboth{IEEE TRANSACTIONS ON MULTIMEDIA,~Vol.~00, No.~0, *~2021}%
{Zhang \MakeLowercase{\textit{et al.}}: Affinity Fusion Graph-based Framework for Natural Image Segmentation}
%



\maketitle

\begin{abstract}
This paper proposes an affinity fusion graph framework to effectively connect different graphs with highly discriminating power and nonlinearity for natural image segmentation. The proposed framework combines adjacency-graphs and kernel spectral clustering based graphs (KSC-graphs) according to a new definition named affinity nodes of multi-scale superpixels. These affinity nodes are selected based on a better affiliation of superpixels, namely subspace-preserving representation which is generated by sparse subspace clustering based on subspace pursuit. Then a KSC-graph is built via a novel kernel spectral clustering to explore the nonlinear relationships among these affinity nodes. Moreover, an adjacency-graph at each scale is constructed, which is further used to update the proposed KSC-graph at affinity nodes. The fusion graph is built across different scales, and it is partitioned to obtain final segmentation result. Experimental results on the Berkeley segmentation dataset and Microsoft Research Cambridge dataset show the superiority of our framework in comparison with the state-of-the-art methods. The code is available at \url{https://github.com/Yangzhangcst/AF-graph}.
\end{abstract}


%
\IEEEpeerreviewmaketitle

\section{Introduction}

\IEEEPARstart{I}{mage} segmentation is a fundamental yet challenging task in computer vision, playing an important role in many practical applications~\cite{8362943}. In the past few years, many image segmentation methods have been developed, which are roughly classified into supervised, semi-supervised, and unsupervised methods. In some scenarios, supervised methods cannot provide a reliable solution when it is difficult to obtain a large number of precisely annotated data. Because no prior knowledge is available, unsupervised image segmentation is a challenging and intrinsically ill-posed problem. Unsupervised methods still receive much attention, because images are segmented without user intervention.

\begin{figure}[!t]
  \centering
  \subfloat[Input image]{
    \includegraphics[width=1in]{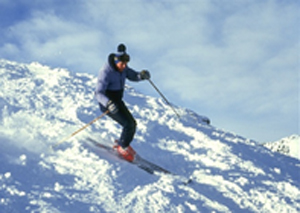}
  }
  \subfloat[Superpixels (s1)]{
    \includegraphics[width=1in]{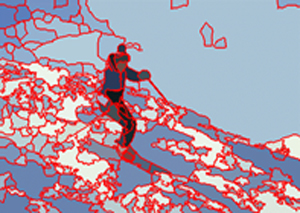}
  }
  \subfloat[Superpixels (s2)]{
    \includegraphics[width=1in]{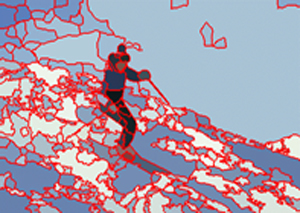}
  }

  \vspace{-0.5em}
  \subfloat[Superpixels (s3)]{
    \includegraphics[width=1in]{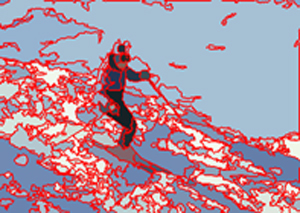}
  }
  \subfloat[Adjacency-graph]{
    \includegraphics[width=1in]{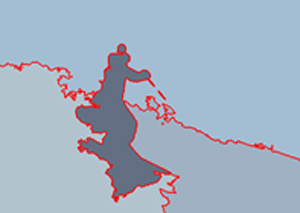}
  }
  \subfloat[$\ell$$_0$-graph]{
    \includegraphics[width=1in]{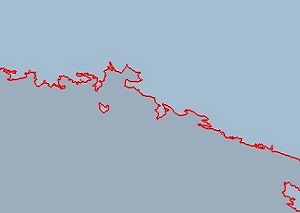}
  }

  \vspace{-0.5em}
  \subfloat[GL-graph]{
    \includegraphics[width=1in]{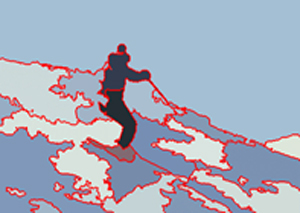}
  }
  \subfloat[AASP-graph]{
    \includegraphics[width=1in]{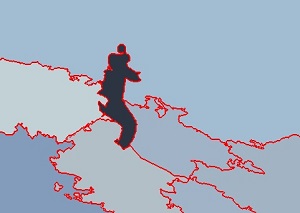}
  }
  \subfloat[AF-graph]{
    \includegraphics[width=1in]{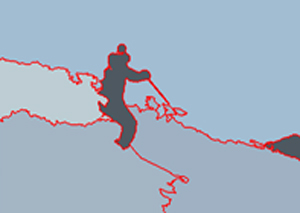}
  }
  \caption{Comparison results by different graph-based segmentation methods. Although superpixel features vary greatly at different scales (s1$\sim$s3),
  our AF-graph can achieve the best performance.}
  \label{fig:graph-based-seg}
\end{figure}

In the literature, many unsupervised segmentation methods have been intensively studied~\cite{Pereyra2017FastUB,Hettiarachchi2016}. Among them, unsupervised graph-based methods usually represent the image content, which have become popular. Because the graphs can embody both spatial and feature information~\cite{LiJSH16}, forming an intermediate image representation for better segmentation. Some representative methods rely on building an affinity graph according to the multi-scale superpixels~\cite{6247750,6738828,7024152,zhang2019aaspgraph}. Especially for affinity graph-based methods, segmentation performance significantly depends on the constructed graph with particular emphasis on the graph topology and pairwise affinities between nodes.

As shown in Fig.~\ref{fig:graph-based-seg}, adjacency-graph~\cite{6247750} usually fails to capture global grouping cues, leading to wrong segmentation results when the objects take up large part of the image. $\ell$$_0$-graph~\cite{6738828} approximates each superpixel with a linear combination of other superpixels, which can capture global grouping cues in a sparse way. But it tends not to emphasize the adjacency, easily incurring isolated regions in segmentation. Both GL-graph~\cite{7024152} and AASP-graph~\cite{zhang2019aaspgraph} combine adjacency-graph with $\ell$$_0$-graph by local and global nodes of superpixels, achieving a better segmentation than a single graph. These nodes are classified adaptively according to superpixel areas in GL-graph and superpixel affinities in AASP-graph. However, there still exist three difficulties to be solved: \textbf{i}) it is not reliable to determine a principle for graph combination, which usually relies on empiricism; \textbf{ii}) local and global nodes are not simply and easily defined because superpixel features change greatly at different scales, especially for superpixel areas; \textbf{iii}) linear graphs, such as $\ell$$_0$-graph, fail to exploit nonlinear structure information of multi-scale superpixels.

To solve these problems, we first need to explore the relationship between different graphs in principle. For graph-based segmentation, it is common to approximate the features of each superpixel using a linear combination of its neighboring superpixels. Such an approximation is regarded as subspace-preserving representation between neighboring superpixels~\cite{7780794}, which is perceived to be the theoretical guidance of graph combination. Because of the sparsity in subspace-preserving representation, its variation is not obvious. Given a subspace-preserving representation of superpixels in each scale, we build an affinity matrix between every pair superpixels (nodes) and apply spectral clustering~\cite{6482137} to select affinity nodes. However, due to the linearity of the representation, segmentation with linear graphs easily result in isolated regions, such as $\ell$$_0$-graph. To enrich the property of fusion graph while improving segmentation performance, it is necessary to capture nonlinear relationships between superpixels. Beside adjacency-graph, we build an affinity graph (KSC-graph) based on spectral clustering in a kernel space~\cite{ZHANG2010959} which is well known for its ability to explore the nonlinear structure information.

In this paper, we propose an affinity fusion graph (\emph{AF-graph}) to integrate adjacency-graph and KSC-graph for natural image segmentation. Our AF-graph combines the above two kinds of graphs based on affinity nodes of multi-scale superpixels with their subspace-preserving representation. The representation is obtained by our proposed sparse subspace clustering based on subspace pursuit (SSC-SP). Furthermore, to discover the nonlinear relationships among the selected nodes, we propose a novel KSC-graph by kernel spectral clustering. And the KSC-graph is further updated upon an adjacency-graph constructed by all superpixels. We evaluate the performance of our AF-graph by conducting experiments on the Berkeley segmentation database and Microsoft Research Cambridge database using four quantitative metrics, including PRI, VoI, GCE, and BDE. The experimental results show the superiority of our AF-graph compared with the state-of-the-art methods.

This work makes the following contributions.

\begin{itemize}
  \item We propose AF-graph to combine different graphs following a novel definition named affinity nodes for natural image segmentation.
  \item The affinity nodes of superpixels are selected according to their subspace-preserving representation through our proposed SSC-SP.
  \item We propose a novel KSC-graph to capture nonlinear relationships of the selected affinity nodes while improving segmentation performance.
\end{itemize}

The rest of this paper is organized as follows. Related works are reviewed in Section II. The proposed AF-graph for natural image segmentation is presented in Section III. Experimental results are reported in Section IV, and the paper is concluded in Section V.

\section{Related Works}
The core of graph-based segmentation is to construct a reliable graph representation of an image. Numerous works have been developed in recent years. These methods can be roughly classified into two categories depending on whether the process of graph construction is unsupervised or not.

\textit{Unsupervised} methods represent the image content with static graphs or adaptive graphs. In static graphs, hard decision is used to select connected nodes, and pairwise similarity is computed without considering other superpixels. An adjacency-graph~\cite{6247750} is built upon each superpixel which is connected to all its neighborhoods. The pairwise similarity is computed by Gaussian kernel function which is influenced easily by the standard deviation~\cite{JShi200,1467569}. In adaptive graphs, pairwise similarity is computed on all data points. Wang \emph{et al.}~\cite{6738828} proposed a $\ell$$_0$-graph by building an affinity graph using $\ell$$_0$ sparse representation. The GL-graph~\cite{7024152} and AASP-graph~\cite{zhang2019aaspgraph} are fused by adjacency-graph~\cite{6247750} and $\ell$$_0$-graph~\cite{6738828}, which can achieve a better result than a single graph. Yin \textit{et al.}~\cite{YIN2017245} utilized a bi-level segmentation operator to achieve multilevel image segmentation, which determines the number of regions automatically. Superpixels with more accurate boundaries are produced in~\cite{Lei2018fuzzy} by a multi-scale morphological gradient reconstruction~\cite{8733206}, and the combination of color histogram and fuzzy c-means achieves fast segmentation. Fu \textit{et al.}~\cite{7410546} relied on contour and color cues to build a robust image segmentation approach, and spectral clustering are combined to obtain great results. A feature driven heuristic four-color labeling method is proposed in~\cite{LiTLL18}, which generates more effective color maps for better results. Cho \textit{et al.}~\cite{7484679} proposed non-iterative mean-shift~\cite{JShi200} based image segmentation using global and local attributes. This method modifies the range kernel in the mean-shift processing to be anisotropic with a region adjacency graph (RAG) for better segmentation performance.

\textit{Semi-supervised/supervised} methods optimize different features and their combinations to measure pairwise affinity for manual image segmentation~\cite{Pablo2011Contour}. These methods also learn a pairwise distance measure using diffusion-based learning~\cite{8453831} and semi-supervised learning~\cite{6341755}. Tighe \textit{et al.}~\cite{TIGHE2010Sup} proposed a superparsing method to compute the class score of superpixels by comparing the nearest neighbor superpixel from retrieval dataset, and infer their labels with Markov random field. Bai \emph{et al.}~\cite{4815272} proposed a method to define context-sensitive similarity. A correlation clustering model is proposed in~\cite{Kim2013Task}, which is trained by support vector machine and achieve task-specific image partitioning. Kim \textit{et al.}~\cite{6727483} considered higher-order relationships between superpixels using a higher-order correlation clustering (HO-CC). Yang \emph{et al.}~\cite{6165308} proposed to learn the similarities on a tensor product graph (TPG) obtained by the tensor product of original graph with itself, taking higher order information into account.

To sum up, the performance of these graph-based segmentation depends on the constructed graph, and the graph is always fixed by neighborhood relationships between superpixels. To address this issue, we propose an adaptive fusion graph-based framework to construct a reliable graph by connecting the adjacency-graph and a novel adaptive-graph of multi-scale superpixels. Inspired by HO-CC~\cite{6727483} and the kernel with RAG~\cite{7484679}, we build an adaptive-graph by kernel spectral clustering to capture the nonlinear relationships between superpixels. The combination of adjacency-graph and adaptive-graph can enrich the property of a fusion graph and further improve the segmentation performance.

\section{Method}

\begin{figure}[!t]
  \centering
  \includegraphics[width=3.4in]{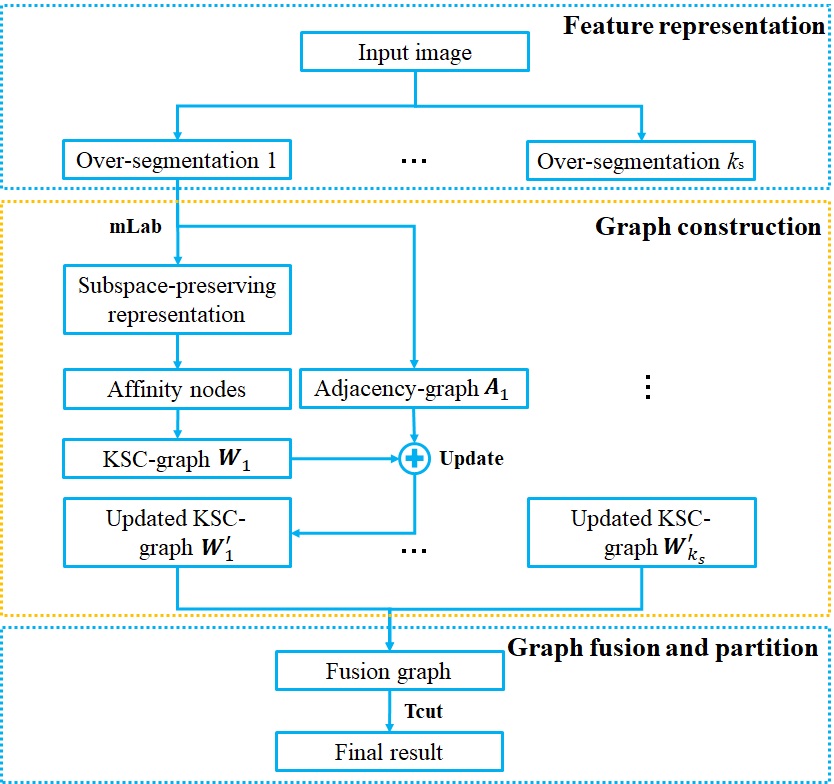}
  \caption{An overview of the proposed AF-graph for natural image segmentation. After over-segmenting an input image, we obtain superpixels with mLab features at different scales. An adjacency-graph is constructed by every superpixel at each scale. To better reflect the affiliation of superpixels at each scale, a subspace-preserving representation of them is obtained to further select affinity nodes. The KSC-graph is built upon the selected nodes, and then updated by adjacency-graph across different scales. The final result is obtained by partitioning the constructed fusion graph through Tcut.}
  \label{fig:pipeline}
\end{figure}

In this section, we introduce our AF-graph for natural image segmentation. The overview of our framework is shown in Fig.~\ref{fig:pipeline}. Our AF-graph primarily consists of three components: feature representation, graph construction, and graph fusion and partition.

\subsection{Feature representation}
The main idea of superpixel generation is grouping similar pixels into perceptually meaningful atomic regions which always conform well to the local image structures. More importantly, superpixels generated by different methods with different parameters can capture diverse and multi-scale visual contents of an image. We simply over-segment an input image into superpixels by mean shift (MS)~\cite{Comaniciu2002} and Felzenszwalb-Huttenlocher (FH) graph-based method~\cite{Felzenszwalb2004Efficient} using the same parameters\footnote{The parameters of oversegmentation have been discussed in~\cite{6247750} and~\cite{6738828}.} (\emph{e.g.} scale $k_s=5$) as done by SAS~\cite{6247750}. Then, the color features of each superpixel are computed to obtain an affinity fusion graph. In our implementation, color feature is formed by mean value in the CIE L*a*b* space (mLab) which can approximate human vision and its L component closely matches the human perception of lightness~\cite{7024152}.
We define the superpixels of an input image $\emph{\textbf{I}}_p$ as $\textbf{\emph{X}}_{k_s}=\{X_i\}^{N}_{i=1}$ at scale $k_s$ with \emph{N} denoting the superpixel numbers. And $\textbf{\emph{F}}=[\ \emph{\textbf{f}}_1,...,\textbf{\emph{f}}_N]\in \mathbb{R}^{n\times N}$ is denoted as mLab feature matrix of the superpixels.

\subsection{Graph construction}
\label{sec:graphconstruction}
For single graph-based image segmentation, a crucial issue is how to approximate each superpixel in the feature space using a linear combination of its neighboring superpixels. Such an approximation is called as subspace-preserving representation between neighboring superpixels, which is calculated from the corresponding representation error~\cite{7024152}. Such a subspace-preserving representation can be formally written as follows:
\begin{equation}
\textbf{\emph{f}}_{j}=\textbf{\emph{F}}\textbf{\emph{c}}_{j}, c_{jj}=0,
\end{equation}
where $\textbf{\emph{c}}_{j}\in\mathbb{R}^N$ is the sparse representation of superpixels, and $\textbf{\emph{f}}_{j}\in\mathbb{R}^n$ over the $\textbf{\emph{F}}$ is a matrix representation of superpixels. The constraint $c_{jj}=0$ prevents the self-representation of $\textbf{\emph{f}}_{j}$.

For fusion graph-based image segmentation, it is critical to generate the subspace-preserving representation for graph combination. Based on the representation of superpixels, an affinity between every pair nodes is built and further selected through spectral clustering~\cite{6482137}. The selected affinity nodes are used to integrate comprehensively different graphs. Moreover, due to the linearity of the subspace-preserving representation, segmentation with a linear graph usually results in isolated regions. In contrast, we propose a KSC-graph to exploit nonlinear structure information. The proposed graph construction consists of three steps: selecting affinity nodes, constructing KSC-graph on affinity nodes, and updating KSC-graph.

\renewcommand{\algorithmicrequire}{\textbf{Input:}}
\renewcommand{\algorithmicensure}{\textbf{Output:}}
\begin{algorithm}[!t]
\caption{Subspace Pursuit (SP)}
\begin{algorithmic}[1]
\REQUIRE $\textbf{\emph{F}}=[\ \textbf{\emph{f}}_1,...,\textbf{\emph{f}}_N]\in \mathbb{R}^{n\times N}$; $\textbf{\emph{b}}\in\mathbb{R}^{n}$; $L_{max}=3$; $\tau=10^{-6}$;
\STATE \textbf{Initialize} $j=0$; residual $\textbf{\emph{r}}_0=\textbf{\emph{b}}$; support set $T_0=\emptyset$;
\WHILE {$\|\textbf{\emph{r}}_{j+1}\|_2-\|\textbf{\emph{r}}_{j}\|_2>\tau$}
    \STATE  $T_t=T_j\bigcup\{i^{*}\}, where\ i^{*}= \mathop{\max}_{j}\{|\textbf{\emph{f}}_{j}^{\mathrm{T}}\textbf{\emph{r}}_j|^1,L_{max}\}$;
    \STATE  $T_{j+1}= \mathop{\max}_{j}\{|\ \textbf{\emph{f}}_{T_t}^{\mathrm{T}}\textbf{\emph{b}}|^1,L_{max}\}$;
    \STATE  $\textbf{\emph{r}}_{j+1}=(I-P_{T_{j+1}})\textbf{\emph{b}}$, where $P_{T_{j+1}}$ is the projection onto the span of vectors $\{\ \textbf{\emph{f}}_j, j\in T_{j+1}\}$;
    \STATE  $j\leftarrow j+1$.
\ENDWHILE
\ENSURE $\textbf{\emph{c}}^{*}=\textbf{\emph{f}}_{T_{j+1}}^{\mathrm{T}}\textbf{\emph{b}}$.
\end{algorithmic}
\end{algorithm}

\begin{algorithm}[!t]
	\caption{Affinity nodes selection (SSC-SP)}
	\begin{algorithmic}[1]
		\REQUIRE $\textbf{\emph{F}}=[\ \textbf{\emph{f}}_1,...,\textbf{\emph{f}}_N]\in \mathbb{R}^{n\times N}$; $L_{max}=3$; $\tau=10^{-6}$;
		\STATE Compute $\textbf{\emph{c}}_j^{*}$ from $SP(\textbf{\emph{F}}_{-j},\textbf{\emph{f}}_j)$;
		\STATE Set $\textbf{\emph{C}}^{*}=[\textbf{\emph{c}}_1^{*},...,\textbf{\emph{c}}_N^{*}]$ and $\textbf{\emph{W}}_{sp}=|\textbf{\emph{C}}^{*}|+|{\textbf{\emph{C}}^{*}}^\mathrm{T}|$;
		\STATE Compute classification from $\textbf{\emph{W}}_{sp}$ by spectral clustering.
		\ENSURE Affinity nodes.
	\end{algorithmic}
\end{algorithm}

\subsubsection{Selecting affinity nodes} As discussed above, affinity nodes are not easily and simply defined because the features of multi-scale superpixels change greatly. Although the features vary greatly, the affiliation of multi-scale superpixels may not be changed. It is proved that the affiliation of superpixels can be well approximated by a union of low-dimensional subspaces~\cite{7780794}. Therefore, affinity nodes can be classified according to the affiliation of superpixels with subspace approximation. This affiliation is considered as subspace-preserving representation which is obtained by sparse subspace clustering (SSC). In the SSC, the subspace clustering is implemented by finding a sparse representation of each superpixel in terms of other superpixels. In principle, we compute it by solving the following optimization problem:
\begin{equation}
\textbf{\emph{c}}_{j}^{*}= \mathop{\arg\min}_{\textbf{\emph{c}}_j}\|\textbf{\emph{c}}_{j}\|_0 \quad s.t.\ \textbf{\emph{f}}_j=\textbf{\emph{F}}\textbf{\emph{c}}_j, \ c_{jj}=0,
\end{equation}
where $\|\cdot\|_0$ represents the $\ell$$_0$-norm. We solve this problem by subspace pursuit (SP)~\cite{4839056} which is summarized in~\textbf{Algorithm 1}. The vector $\textbf{\emph{c}}_j^{*}\in \mathbb{R}^{N}$ (the \emph{j}-th column of $\textbf{\emph{C}}^{*}\in\mathbb{R}^{N\times N}$) is computed by $SP(\textbf{\emph{F}}_{-j}, \textbf{\emph{f}}_j)\in \mathbb{R}^{N-1}$ with a zero inserted in its \emph{j}-th entry, where $\textbf{\emph{F}}_{-j}$ is the feature matrix with the \emph{j}-th column removed. The SP is subject to a complexity of $O(NL_{max})$.

\begin{figure}[!t]
  \centering    
    \subfloat[Superpixels]{
    	\includegraphics[height=3in]{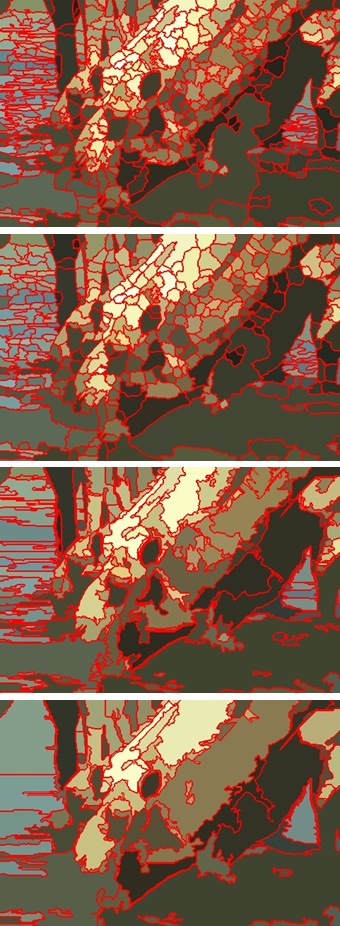}
    }\hspace{-0.8em}
    \subfloat[SRs]{
    	\includegraphics[height=3in]{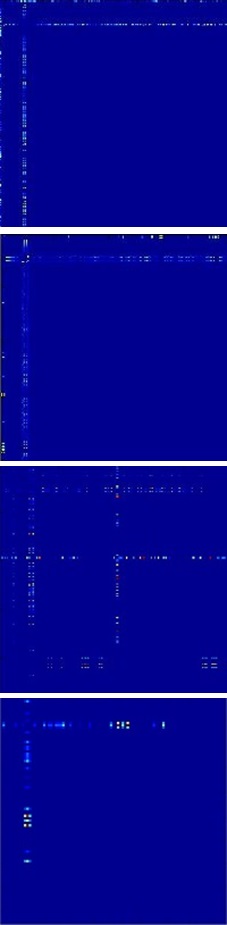}
    }\hspace{-0.8em}   
    \subfloat[Areas]{
    	\includegraphics[height=3in]{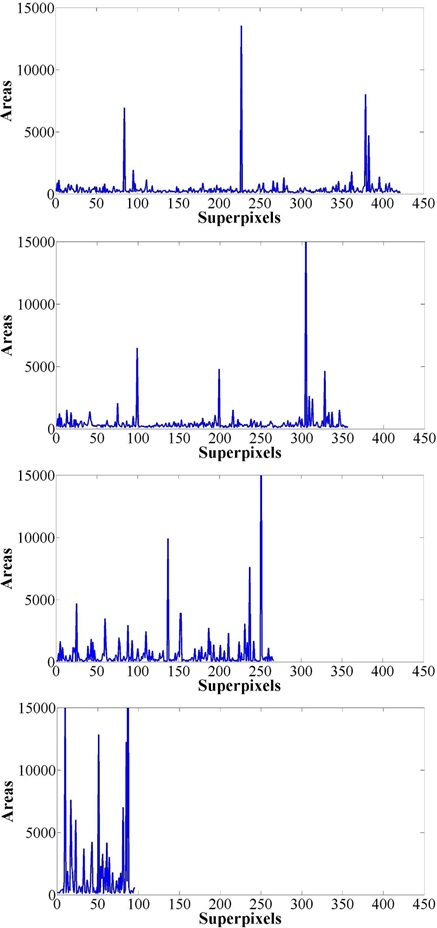}
    }
  \caption{Illustration of subspace-preserving representations (SRs) and areas computed by  superpixels across different scales. The subspace-preserving representations of multi-scale superpixels are always sparse. From top to bottom, the variation of the representations is obviously smaller than that of areas with the decreasing number of superpixels. Compared with the area, the subspace-preserving representation can better reveal the affiliation of superpixels.}
  \label{fig:SPR}
\end{figure}

As shown in Fig.~\ref{fig:SPR}, we compare the presentations generated by multi-scale superpixels with their areas across different scales. It can be seen that the subspace-preserving representations of multi-scale superpixels are always sparse. The variation of presentations is obviously smaller than that of areas with the decreasing number of superpixels. Compared with the area, subspace-preserving representation can better reveal the affiliation of superpixels. Based on the presentations, the classification of superpixels is obtained by applying our proposed SSC-SP. The procedure of affinity node selection (SSC-SP) is summarized in~\textbf{Algorithm 2}. 

\subsubsection{Constructing KSC-graph on affinity nodes} In general, the basic principle of $\ell$$_0$ sparse representation for graph construction is that each superpixel is approximated with a linear combination of other superpixels. Therefore, $\ell$$_0$-graph cannot exploit nonlinear structure information of superpixels. To discover nonlinear relationships among the selected affinity nodes, we consider the following problem:
\begin{equation}
 \mathop{\min}_{\textbf{\emph{Z}}}\|\textbf{\emph{X}}-\textbf{\emph{XZ}}\|_F^2+\alpha\|\textbf{\emph{Z}}\|_1  \ s.t.\ \textbf{\emph{Z}}^{\mathrm{T}}\textbf{1}=1, \ 0\leq z_{ij}\leq1,
\end{equation}
where $\|\cdot\|_F^2$ represents the squared Frobenius norm, $z_{ij}$ is the $(i,j)$-th element of similarity graph matrix \emph{\textbf{Z}}, and $\alpha>0$ is a balance parameter.

It is recognized that nonlinear data can represent linearity when mapped to an implicit, higher-dimensional space via a kernel function~\cite{KangPC17}. All similarities among the superpixels can be computed exclusively using the kernel function, and the transformation of superpixels does not need to be known. Such approach is well-known as the kernel trick, and it greatly simplifies the computation in kernel space~\cite{ZHANG2010959} when a kernel is precomputed. To fully exploit nonlinearity of superpixels, we define $\phi: \mathcal{R}^\textbf{\emph{D}}\rightarrow\mathcal{H}$ to be a function mapping superpixels from an input space to a reproducing kernel Hilbert space $\mathcal{H}$, where $\textbf{\emph{D}}\in \mathbb{R}^{N\times N}$ is a diagonal matrix with the \emph{i}-th diagonal element $\sum_j\frac{1}{2}(z_{ij}+z_{ji})$. The transformation of superpixels $\textbf{\emph{X}}$ at a certain scale is $\phi(\textbf{\emph{X}})=\{\phi(X_i)\}^{N}_{i=1}$. The kernel similarity between superpixels $X_i$ and $X_j$ is defined by a predefined kernel as $K_{X_i,X_j}=<\phi(X_i),\phi(X_j)>$. In practice, we use a linear kernel or Gaussian kernel in our experiments. 

This model recovers the linear relationships between superpixels in the space $\mathcal{H}$ and thus the nonlinear relations in original representation~\cite{KangPC17}. So, the problem is formulated as:
\begin{equation}
\begin{aligned}
\mathop{\min}_{\textbf{\emph{Z}}}Tr&(\textbf{\emph{K}}-2\textbf{\emph{KZ}}+\textbf{\emph{Z}}^\mathrm{T}\textbf{\emph{KZ}})+\alpha\|\textbf{\emph{Z}}\|_1\!+\!\beta Tr(\textbf{\emph{P}}^\mathrm{T}\textbf{\emph{LP}}) \\
&s.t.\ \textbf{\emph{Z}}^{\mathrm{T}}\textbf{1}=1, \ 0\leq z_{ij}\leq1,\ \textbf{\emph{P}}^\mathrm{T}\textbf{\emph{P}}=\textbf{\emph{I}},
\end{aligned}
\label{eq_ksc}
\end{equation}
where $Tr(\cdot)$ is the trace operator, and $\textbf{\emph{L}}$ is the Laplacian matrix. $\beta>0$ is also a balance parameter. $\textbf{\emph{P}}\in \mathbb{R}^{N\times c}$ is the indicator matrix. The \emph{c} elements of \emph{i}-th row $\textbf{\emph{P}}_{i,:}\in \mathbb{R}^{1\times c}$ are used to measure the membership of superpixel $X_i$ belonging to \emph{c} clusters.

We obtain the optimal solution \emph{\textbf{P}} by \emph{c} eigenvectors of \emph{\textbf{L}} corresponding to \emph{c} smallest eigenvalues. When \emph{\textbf{P}} is fixed, Eq. (\ref{eq_ksc}) is reformulated column-wisely as:
\begin{equation}
\begin{aligned}
\mathop{\min}_{\textbf{\emph{Z}}}K_{ii}-2&\textbf{\emph{K}}_{i,:}\textbf{\emph{Z}}_i+\textbf{\emph{Z}}_i^{\mathrm{T}}\textbf{\emph{KZ}}_i+\alpha \textbf{\emph{Z}}_i^{\mathrm{T}}\textbf{\emph{Z}}_i+ \frac{\beta}{2}\textbf{\emph{e}}_i^{\mathrm{T}}\textbf{\emph{Z}}_i \\
& s.t.\ \textbf{\emph{Z}}^{\mathrm{T}}_i\textbf{1}=1, \ 0\leq z_{ij}\leq1,
\end{aligned}
\label{eq_kscgraph}
\end{equation}
where $\textbf{\emph{e}}_i\in {\mathbb{R}^{N\times1}}$ is a vector with the \emph{j}-th element $e_{ij}$ being $e_{ij}=\|\textbf{\emph{P}}_{i,:}-\textbf{\emph{P}}_{j,:}\|_2$.
This problem can be solved by many existing quadratic programming in polynomial time. The main computation cost lies in solving \emph{\textbf{Z}} which is generally solved in parallel. After solving the above problem, we obtain a symmetric non-negative similarity matrix \emph{\textbf{Z}} to construct KSC-graph as follows: 
\begin{equation}
\emph{\textbf{W}}=(|\textbf{\emph{Z}}|+|{\textbf{\emph{Z}}}|^\mathrm{T})/2.
\end{equation} 
The KSC-graph construction is outlined in~\textbf{Algorithm 3}.

\subsubsection{Updating KSC-graph} Inspired by GL-graph~\cite{7024152}, the local neighborhood relationship of superpixels is also considered to enrich the property of fusion graph and further improve segmentation accuracy. As for all superpixels at each scale, everyone is connected to its adjacent superpixels, denoted as adjacency-graph \emph{\textbf{A}}. Let $\textbf{\emph{\textbf{M}}}_A$ be the matrix-representation of all adjacent neighbors of every superpixel, we attempt to represent the $\textbf{\emph{f}}_i$ as a linear combination of elements in $\textbf{\emph{M}}_A$. In practice, we solve the following optimization problem:
\begin{equation}
\tilde{\textbf{\emph{c}}_{i}}= \mathop{\arg\min}_{\textbf{\emph{c}}_{i}}\|\ \textbf{\emph{f}}_i-\textbf{\emph{M}}_A\textbf{\emph{c}}_i\|_2.
\end{equation}
If a minimizer $\tilde{\textbf{\emph{c}}_{i}}$ has been obtained, the affinity coefficients $A_{ij}$ between superpixels $X_i$ and $X_j$ are computed as $A_{ij}=1-\frac{1}{2}(r_{i,j}+r_{j,i})$ with $r_{i,j}= \|\ \textbf{\emph{f}}_i-\textbf{\emph{c}}_{i,j}\textbf{\emph{f}}_j\|_2^2$, if \emph{i} is not equal to \emph{j}; $A_{ij}=1$ otherwise.
For the superpixels at each scale, the KSC-graph \emph{\textbf{W}} is used to replace the adjacency-graph \emph{\textbf{A}} to obtain the updated KSC-graph $\textbf{\emph{W}}'$ on affinity nodes. 

\begin{figure}[!t]
	\centering
	\includegraphics[width=3.4in]{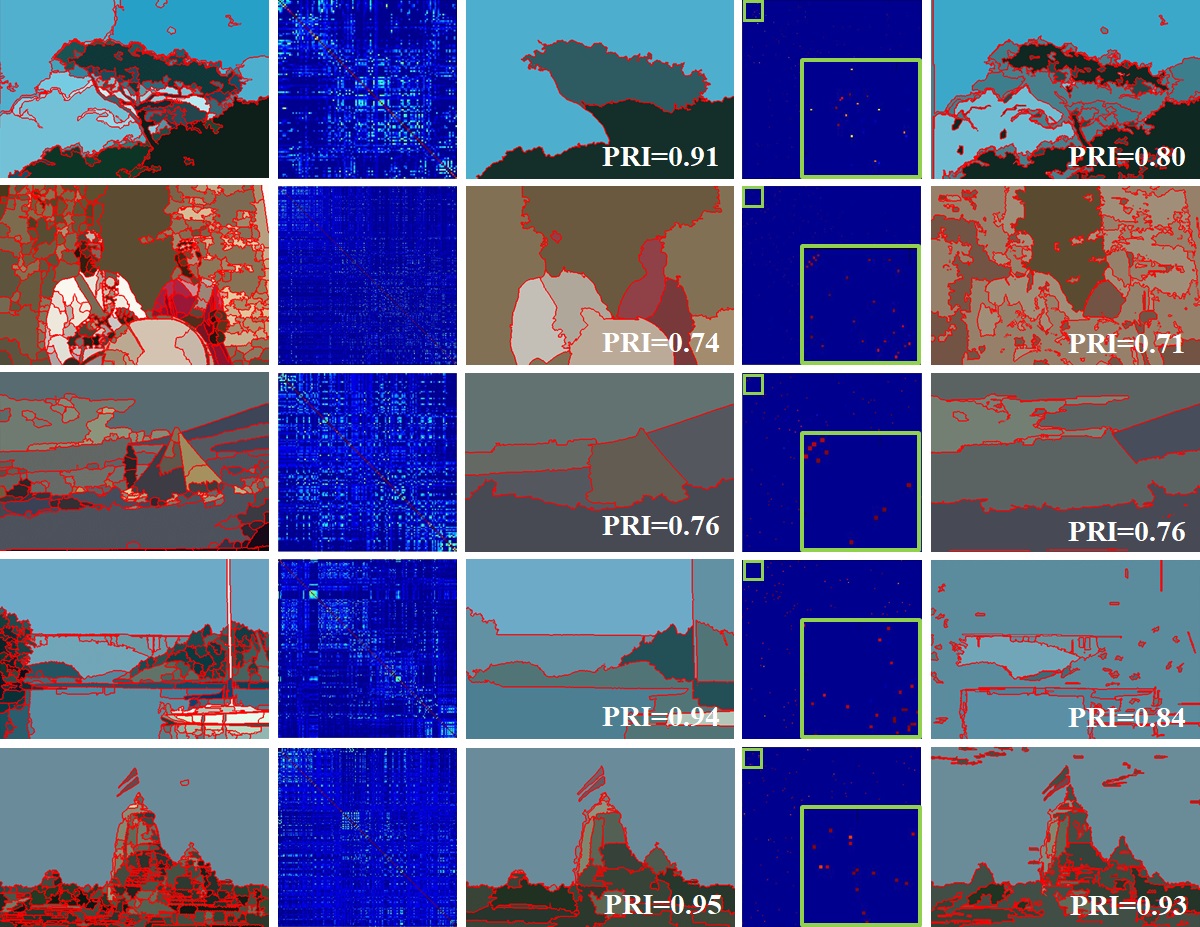}
	\caption{Visual comparison obtained by KSC-graph and $\ell$$_0$-graph. From left to right, superpixel images, KSC-graphs built by the superpixels, segmentation results by the KSC-graphs, $\ell$$_0$-graphs built by the superpixels, segmentation results by the $\ell$$_0$-graphs are presented, respectively. The KSC-graph produces a dense graph, and the $\ell$$_0$-graph is sparser than KSC-graph. The proposed KSC-graph achieves better performance than $\ell$$_0$-graph.}
	\label{fig:adj_ksc}
\end{figure}

To further illustrate the differences between our KSC-graph and $\ell$$_0$-graph, we use probabilistic rand index (PRI)~\cite{4160946} to evaluate the results of KSC-graph and $\ell$$_0$-graph on Berkeley Segmentation Database~\cite{Martin2001A}. As shown in Fig.~\ref{fig:adj_ksc}, our proposed KSC-graph produces a dense graph, and the $\ell$$_0$-graph is sparser than KSC-graph because of its nodes owning fewer neighbors. The proposed KSC-graph achieves better performance in comparison with $\ell$$_0$-graph.

\begin{algorithm}[!t]
	\caption{KSC-graph construction}
	\begin{algorithmic}[1]
		\REQUIRE Kernel matrix \emph{\textbf{K}}; $\alpha>0$; $\beta>0$; $\delta=10^{-3}$;
		\STATE \textbf{Initialize} random matrix \emph{\textbf{Z}} and \emph{\textbf{P}}; $j=0$;
		\WHILE {$\frac{\|\textbf{\emph{Z}}_{j+1}-\textbf{\emph{Z}}_{j}\|_2}{\|\textbf{\emph{Z}}_{j}\|_2}<\delta$}
		\STATE Update \emph{\textbf{P}}, which is formed by the \emph{c} eigenvectors of $\textbf{\emph{L}}=\textbf{\emph{D}}-\frac{\textbf{\emph{Z}}_j^{\mathrm{T}}+\textbf{\emph{Z}}_j}{2}$ corresponding to the \emph{c} smallest eigenvectors;
		\STATE For each \emph{j}, update the \emph{j}-th column of $\textbf{\emph{Z}}_j$ by solving the problem Eq. (\ref{eq_kscgraph}).
		\ENDWHILE
		\STATE Construct KSC-graph $\emph{\textbf{W}}=(|\textbf{\emph{Z}}_{j+1}|+|{\textbf{\emph{Z}}}_{j+1}|^\mathrm{T})/2$
		\ENSURE KSC-graph $\emph{\textbf{W}}$.
	\end{algorithmic}
\end{algorithm}

\begin{algorithm}[!t]
	\caption{Affinity fusion graph-based framework for natural image segmentation (AF-graph)}
	\begin{algorithmic}[1]
		\REQUIRE Input image $\emph{\textbf{I}}_p$; parameters $\alpha$, $\beta$; group $k_T$;
		\STATE Over-segment an input image $\emph{\textbf{I}}_p$ to obtain superpixels at different scales;
		\STATE Generate a subspace-preserving representation of color features at each scale to better represent the superpixels based on the proposed SSC-SP;
		\STATE Select affinity nodes of superpixels based on the proposed subspace-preserving representation;
		\STATE Construct a KSC-graph on the selected nodes through kernel spectral clustering;
		\STATE Construct an adjacency-graph by all superpixels and update the KSC-graph at each scale;
		\STATE Fuse the updated KSC-graph across different scales and compute to obtain the final segmentation result (pixel-wise labels) through Tcut with group $k_T$;
		\ENSURE Pixel-wise labels.
	\end{algorithmic}
\end{algorithm}

\subsection{Graph fusion and partition}
\label{sec:graphpartition}
To fuse all scales of superpixels, we plug each scale affinity matrix $\textbf{\emph{W}}'_{k_s}$ corresponding to its graph into a block diagonal multi-scale affinity matrix $\textbf{\emph{W}}_{MS}$ as follows:

\begin{equation}
\textbf{\emph{W}}_{MS}=
{\begin{pmatrix}
	\textbf{\emph{W}}_1^{'} & \cdots & 0\\
	\vdots & \ddots & \vdots\\
	0 & \cdots & \textbf{\emph{W}}_{k_s}^{'}
	\end{pmatrix}}.
\end{equation}
We construct a fusion graph to describe the relationships of pixels to superpixels and superpixels to superpixels, which aims to enable propagation of grouping cues across superpixels at different scales.  Formally, let $\textbf{\emph{G}}=\{\textbf{\emph{U}},\textbf{\emph{V}},\textbf{\emph{B}}\}$ denote the fusion graph with node set $\textbf{\emph{U}}\cup \textbf{\emph{V}}$, where $\textbf{\emph{U}}:=\textbf{\emph{I}}_p\cup \textbf{\emph{X}}=\{u_i\}_{i=1}^{N_U}$, $\textbf{\emph{V}}:=\textbf{\emph{X}}=\{v_i\}_{i=1}^{N_V}$ with $N_U=|\textbf{\emph{I}}_p|+|\textbf{\emph{X}}|$ and $N_V=|\textbf{\emph{X}}|$, the numbers of nodes in \emph{\textbf{U}} and \emph{\textbf{V}}, respectively. The across-affinity matrix is defined as $\textbf{\emph{B}}=\begin{bmatrix} \textbf{\emph{A}}_X\\  \textbf{\emph{W}}_{MS}\end{bmatrix}$, where $\textbf{\emph{W}}_{MS}$ is the above multi-scale affinity matrix. $\textbf{\emph{A}}_X=(a_{ij})_{|\textbf{\emph{I}}_p|\times |\textbf{\emph{X}}|}$ are the relationships between pixels and superpixels with $a_{ij}=0.001$, if a pixel \emph{i} belongs to a superpixel \emph{j}; $a_{ij}=0$ otherwise. 

For the above fusion graph $\textbf{\emph{G}}$, the task is to partition it into $k_T$ groups. The $k_T$ is a hyper-parameter, which will be further analyzed in Section~\ref{sec:robustness_anal}. However, in this case, the fusion graph is unbalanced (\emph{i.e.} $N_U=N_V+|\textbf{\emph{I}}_p|$, and $|\textbf{\emph{I}}_p|>>N_V$. We have $N_U>>N_V$). We can apply Transfer cuts (Tcut) algorithm~\cite{6247750} to solve this unbalanced fusion graph. It should be noted that solving the problem takes a complexity of $O(k_T|N_V|^{3/2})$ with a constant. The proposed affinity fusion graph-based framework for natural image segmentation is summarized in \textbf{Algorithm 4}.

\section{Experiments and analysis}
In this section, we first introduce datasets and evaluation metrics. Then, we show performance analysis about affinity nodes selection, kernel functions, different graph combination, and parameters. Moreover, the results are presented in comparison with the state-of-the-art methods. Finally, we present time complexity analysis of our AF-graph.

\subsection{Datasets and evaluation metrics}
For natural image segmentation, we evaluate the performance of our framework on Berkeley segmentation database (BSD)~\cite{Martin2001A} and Microsoft Research Cambridge (MSRC) database~\cite{Shotton2006TextonBoost}. The BSD300 includes 300 images and the ground truth annotations. As an improved version of BSD300, the BSD500 contains 500 natural images. The MSRC contains 591 images and 23 object classes with accurate pixel-wise labeled images. Besides, there are four standard metrics which are commonly used for evaluating image segmentation methods: probabilistic rand index (PRI)~\cite{4160946}, variation of information (VoI)~\cite{Meila2007}, global consistency error (GCE)~\cite{Martin2001A}, and boundary displacement error (BDE)~\cite{Freixenet2002Yet}. Among these metrics, the closer the segmentation result is to the ground-truth, the higher the PRI is, and the smaller the other three measures are. Four metrics of PRI, VoI, GCE, and BDE are computed as the method in~\cite{zhang2019aaspgraph}. In addition, the average running time are obtained to evaluate the time complexity of a segmentation method. In practice, the best segmentation results are computed over the value of group $k_T$ ranking from 1 to 40 as done by GL-graph~\cite{7024152}.

\begin{figure*}[!t]
	\centering
	\includegraphics[width=6.9in]{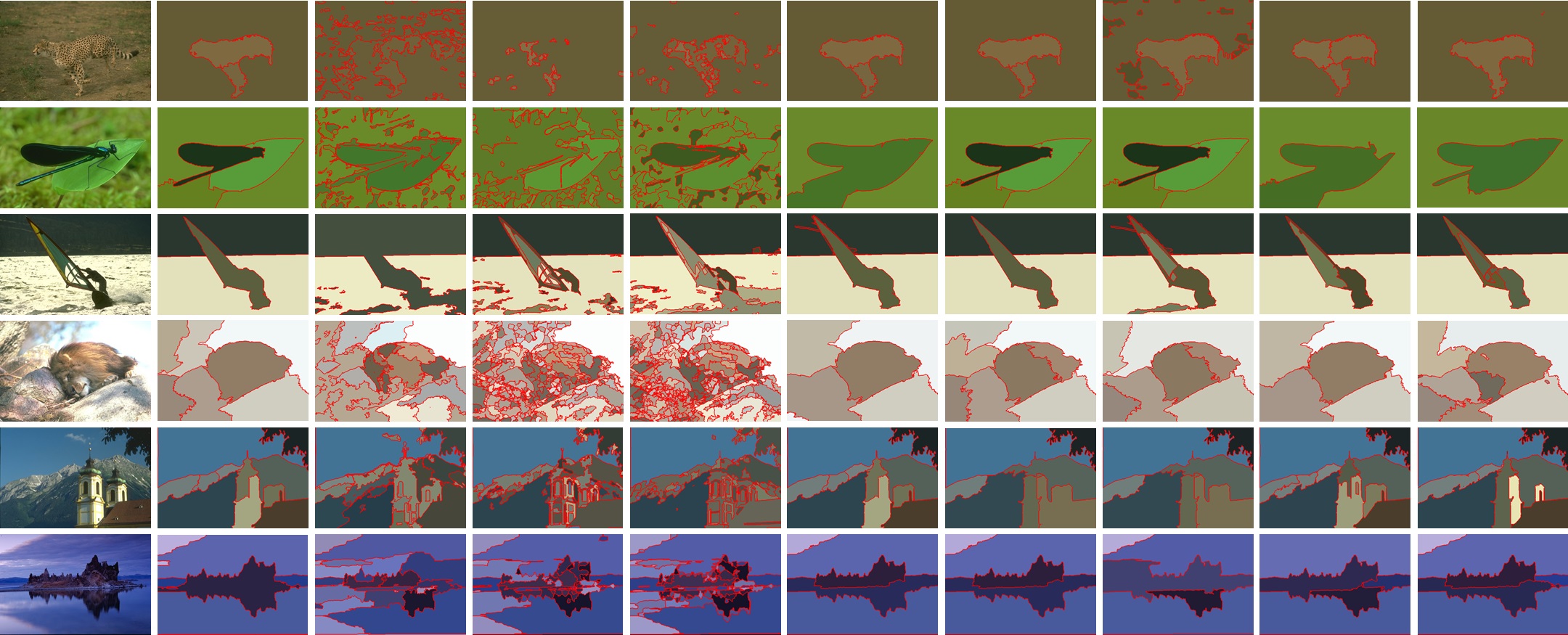}
	\caption{Visual segmentation examples are obtained by basic graphs in comparison with our proposed AF-graph. From left to right, input images, the segmentation results of A-graph, $\ell$$_0$-graph, $\ell$$_1$-graph, $\ell$$_2$-graph, KSC-graph, A+$\ell$$_0$-graph, A+$\ell$$_1$-graph, A+$\ell$$_2$-graph, and our AF-graph (A+KSC-graph) are presented respectively. Our framework can significantly improve the performance of $\ell$$_0$-graph, $\ell$$_1$-graph, and $\ell$$_2$-graph. Our AF-graph can achieve the best performance for the combination of adjacency-graph and KSC-graph.}
	\label{visualper}
\end{figure*}

\begin{figure*}[!t]
	\centering
	\subfloat[PRI]{
		\includegraphics[width=3.1in]{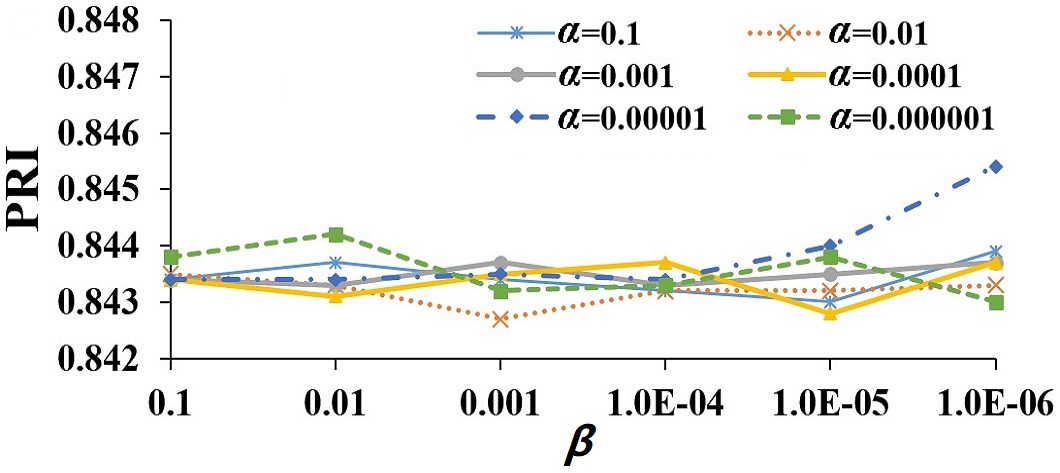}
	}\qquad
	\subfloat[VoI]{
		\includegraphics[width=3.1in]{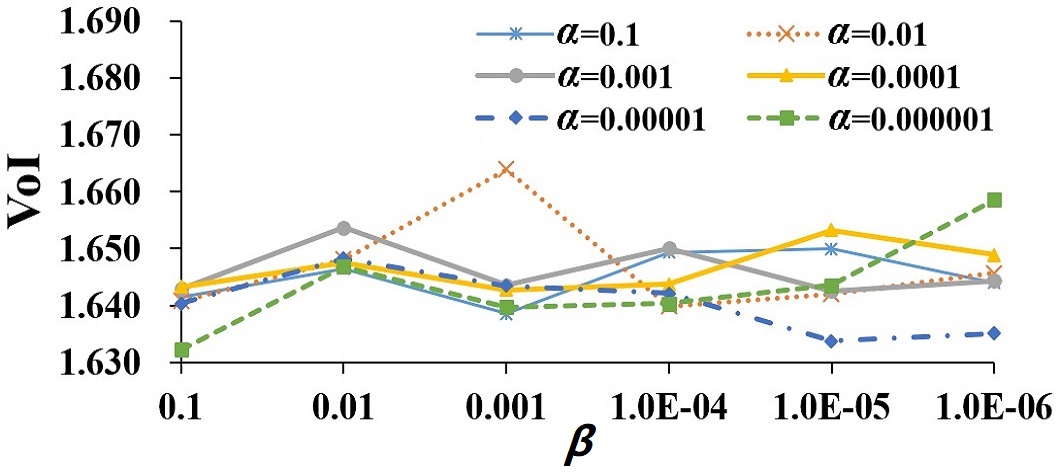}
	}
	
	\vspace{-0.5em}
	\subfloat[GCE]{
		\includegraphics[width=3.1in]{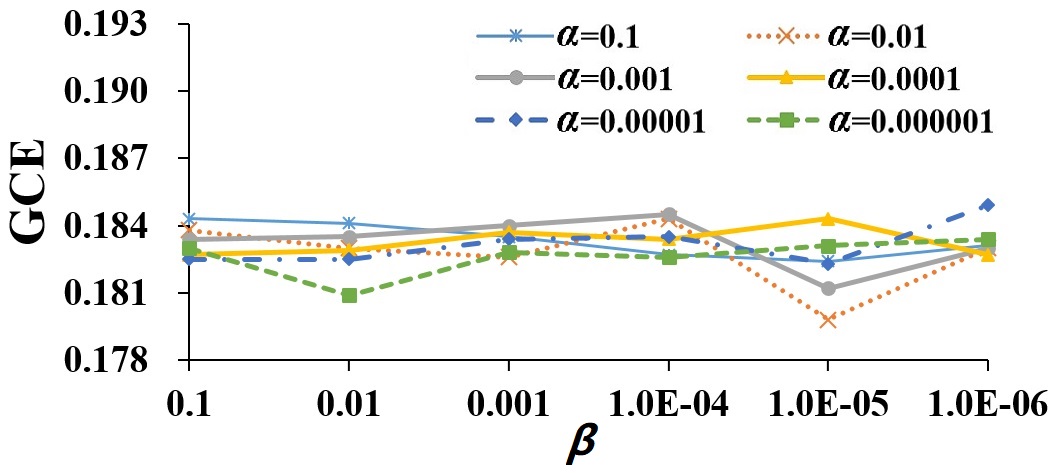}
	}\qquad
	\subfloat[BDE]{
		\includegraphics[width=3.1in]{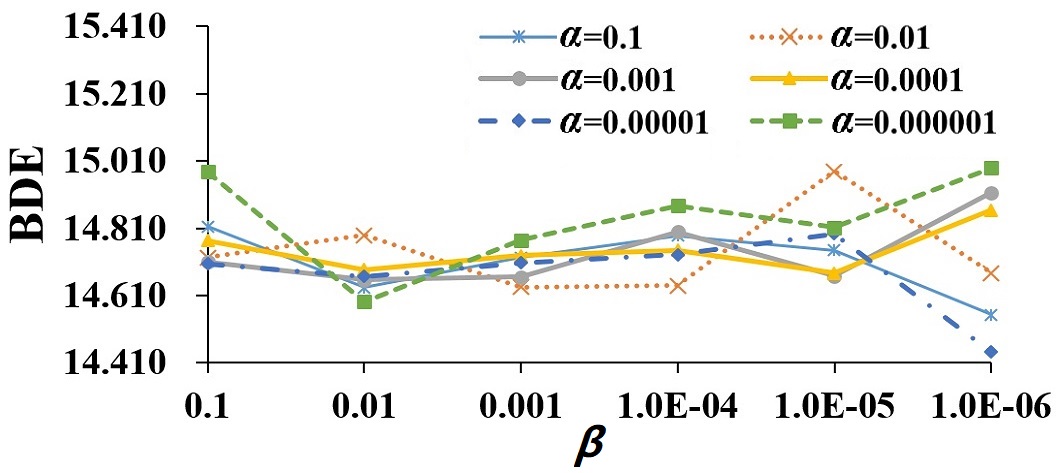}
	}
	\caption{Parameter influence. Each figure is a metric plot when one of the two parameters $\alpha$ and $\beta$ is fixed. The best performance is obtained when $\alpha=10^{-5}$ and $\beta=10^{-6}$. Our AF-graph is robust to the parameters $\alpha$ and $\beta$.}
	\label{fig:parametersin}
\end{figure*}

\begin{figure*}[!t]
	\centering
	\includegraphics[width=6.7in]{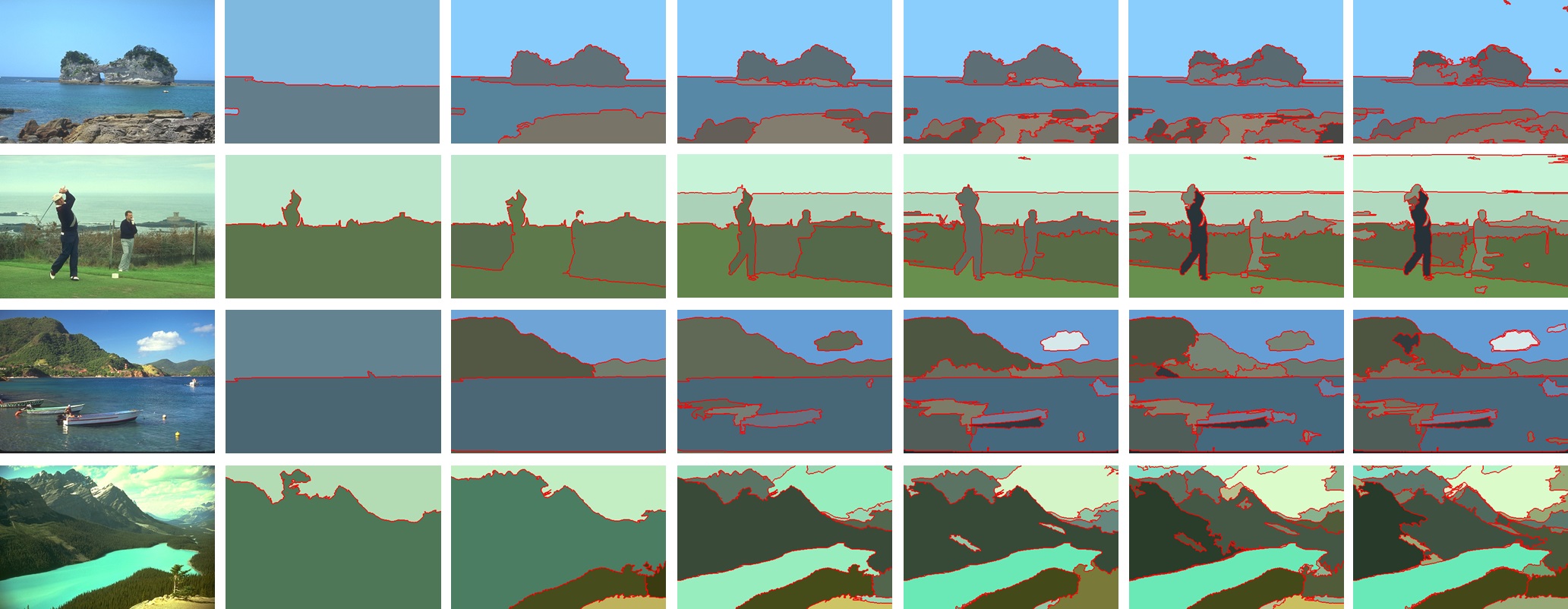}
	\caption{Visual results are obtained by our AF-graph to explore the influence of $k_T$. From left to right, input images, results of $k_T=2$, $k_T=5$, $k_T=10$, $k_T=20$, $k_T=30$, $k_T=40$ are presented respectively. When the $k_T$ is increased, our AF-graph enforces the global structure over superpixels and masters the meaning of regions. Moreover, our AF-graph preserves more local information in superpixels.}
	\label{fig:ktinfluence}
\end{figure*}

\begin{table}[!t]
	\renewcommand{\arraystretch}{1.2}
	\caption{Quantitative comparison on BSD300 dataset for different methods using in the proposed framework. The method in GL-graph~\cite{7024152}, \textit{k}-means, SSC-OMP, and SSC-SP are used for selecting affinity nodes in graph construction.}
	\label{tab:affinity}
	\centering
	\begin{tabular}{lccccc}
		\toprule
		Methods & & PRI $\uparrow$  & VoI $\downarrow$ & GCE $\downarrow$ & BDE $\downarrow$  \\
		\midrule
	    Ref.~\cite{7024152}           & & 0.84 & 1.79 & 0.19 & 15.04  \\		
		\textit{k}-means ($k=2$)  & & 0.84 & 1.69 & 0.19 & 14.72  \\
		SSC-OMP                   & & 0.84 & 1.65 & 0.19 & 14.81  \\	
		SSC-SP                    & & 0.85 & 1.63 & 0.18 & 13.95  \\	
		\bottomrule
	\end{tabular}
\end{table}

\begin{table}[!t]
	\renewcommand{\arraystretch}{1.2}
	\caption{Quantitative comparison on BSD300 dataset for different kernel functions using in our framework.}
	\label{tab:kernel}
	\centering
	\begin{tabular}{lcccc}
		\toprule
		Kernels & PRI $\uparrow$  & VoI $\downarrow$ & GCE $\downarrow$ & BDE $\downarrow$  \\
		\midrule
		Linear                  & 0.85 & 1.63 & 0.19 & 13.95 \\		
		Polynomial $(a=0, b=2)$ & 0.85 & 1.64 & 0.18 & 14.70 \\
		Polynomial $(a=0, b=4)$ & 0.84 & 1.64 & 0.18 & 14.88 \\
		Polynomial $(a=1, b=2)$ & 0.84 & 1.64 & 0.18 & 14.92 \\
        Polynomial $(a=1, b=4)$ & 0.85 & 1.64 & 0.18 & 14.57 \\
		Gaussian $(t=0.1)$ & 0.84 & 1.65 & 0.19 & 14.90 \\	
		Gaussian $(t=1)$   & 0.84 & 1.68 & 0.19 & 14.84 \\	
		Gaussian $(t=10)$  & 0.85 & 1.63 & 0.18 & 14.76 \\	
        Gaussian $(t=100)$ & 0.85 & 1.63 & 0.18 & 13.95 \\	
		\bottomrule
	\end{tabular}
\end{table}

\begin{table}[!t]
	\renewcommand{\arraystretch}{1.2}
	\caption{Performance of different graphs before and after embedding into the proposed framework on BSD300 dataset. The A-graph means adjacency-graph.}
	\label{tab:embed}
	\centering
	\begin{tabular}{llcccc}
		\toprule
		\multicolumn{2}{l}{Graphs} & PRI $\uparrow$  & VoI $\downarrow$ & GCE $\downarrow$ & BDE $\downarrow$ \\
		\midrule
		\multirow{6}{*}{\rotatebox{90}{Not embedded}}
		&A-graph              & 0.84 & 1.64 & 0.18 & 14.73 \\
		&$\ell$$_0$-graph     & 0.83 & 2.08 & 0.23 & 14.96 \\
		&$\ell$$_1$-graph     & 0.80 & 2.96 & 0.33 & 16.08 \\
		&$\ell$$_2$-graph     & 0.80 & 3.03 & 0.32 & 16.33 \\
		&LRR-graph            & 0.84 & 1.74 & 0.20 & 15.27 \\
		&KSC-graph            & 0.84 & 1.74 & 0.20 & 15.41 \\
		\midrule
		\multirow{5.2}{*}{\rotatebox{90}{Embedded}}
		&A + $\ell$$_0$-graph     & 0.84 & 1.66 & 0.19 & 14.69 \\
		&A + $\ell$$_1$-graph     & 0.84 & 1.67 & 0.19 & 14.86 \\
		&A + $\ell$$_2$-graph     & 0.84 & 1.69 & 0.19 & 14.62 \\
		&A + LRR-graph            & 0.84 & 1.64 & 0.18 & 14.76 \\
		&A + KSC-graph (ours)     & 0.85 & 1.63 & 0.18 & 13.95 \\
		\bottomrule
	\end{tabular}
\end{table}

\subsection{Performance analysis}
\label{sec:robustness_anal}

\textbf{Selecting affinity nodes.}
To see how the affinity node selection is affected by different methods, we compare the method in GL-graph~\cite{7024152}, \textit{k}-means ($k=2$), and SSC-OMP~\cite{7780794} with our proposed SSC-SP. All parameters of the above algorithms are set to default values. The comparison results on BSD300 dataset are shown in Table~\ref{tab:affinity}. Obviously, our proposed SSC-SP performs the best on all metrics compared with the method in GL-graph~\cite{7024152}, \textit{k}-means, and SSC-OMP~\cite{7780794}. Therefore, our subspace-preserving representation which is generated by SSC-SP can better reveal the affiliation of multi-scale superpixels.

\textbf{Kernel function in KSC-graph.}
To assess the effectiveness of different kernel functions in KSC-graph, we adopted 9 different kernels including a linear kernel $\textbf{\emph{K}}(\textbf{\emph{x}}, \textbf{\emph{y}})=\textbf{\emph{x}}^{\mathrm{T}}\textbf{\emph{y}}$, four polynomial kernels $\textbf{\emph{K}}(\textbf{\emph{x}}, \textbf{\emph{y}})=(a + \textbf{\emph{x}}^{\mathrm{T}}\textbf{\emph{y}})^b$ with $a\in\{0,1\}$ and $b\in\{2,4\}$, and four Gaussian kernels $\textbf{\emph{K}}(\textbf{\emph{x}}, \textbf{\emph{y}})=exp(-\|\textbf{\emph{x}}-\textbf{\emph{y}}\|_2^2/(td_{max}^2))$ with $d_{max}$ being the maximal distance between superpixels and $t$ varying in the set $\{0.1, 1, 10, 100\}$. Furthermore, all kernels are rescaled to [0, 1] by dividing each element by the largest pair-wise squared distance. In Table~\ref{tab:kernel}, the Gaussian kernel with $t=100$ achieves the best performance on all metrics. We can also observe that our AF-graph is robust to the kernel functions. Specially, when simply using a linear kernel, our framework can still achieve a satisfactory performance.

\textbf{Combining different graphs.}
We construct different basic graphs only using mLab to obtain segmentation results on BSD300. They are adjacency-graph~\cite{6247750}, $\ell$$_0$-graph~\cite{6738828}, $\ell$$_1$-graph~\cite{6482137}, $\ell$$_2$-graph~\cite{7434007}, LRR-graph~\cite{6180173}, and our proposed KSC-graph. Then, we employ these basic graphs to construct a fusion graph. The parameters in each of these graphs are tuned for the best performance. The segmentation results of the above graphs are shown in Table~\ref{tab:embed}.

For basic graphs, many works on linear graph partitioning show that meaningful results are derived from a sparse graph such as $\ell$$_0$-graph. The adjacency-graph shows better performance for all metrics compared with all single graphs. The main reason is that the LRR-graph often produces a dense graph. The $\ell$$_0$-graph performs better than the $\ell$$_1$-graph and $\ell$$_2$-graph. Because the $\ell$$_0$-graph is sparser than the $\ell$$_1$-graph and $\ell$$_2$-graph due to the fewer neighbors of its nodes. So, the sparsity of a linear graph has a great influence on its segmentation performance. We can also find that our KSC-graph produces a dense graph (in Fig.~\ref{fig:adj_ksc}), but its performance is desirable because of the non-linearity. Our AF-graph is more precise for graph combination with respect to all metrics due to assimilating the advantages of different graphs.

More importantly, our AF-graph can significantly improve the performance of these graphs. The metric of PRI has been enhanced largely and the other three metrics have been greatly reduced. Some visual segmentation examples of our AF-graph in comparison with the basic graphs before and after embedded into our framework are shown in Fig.~\ref{visualper}. Clearly, our AF-graph can achieve the best performance.

\textbf{Parameters.} To analyze the robustness of our framework, we study the sensitivity of the two parameters $\alpha$ and $\beta$ in KSC-graph by fixing one of them to the optimal settings. Parameter influence on metrics of the KSC-graph is shown in Fig.~\ref{fig:parametersin}. The best performance is obtained when $\alpha=10^{-5}$ and $\beta=10^{-6}$. We can also observe that our AF-graph is robust to the parameters $\alpha$ and $\beta$.

To explore the influence of group $k_T$ to our AF-graph, we show visual results of various $k_T$ ($k_T$ = 2, 5, 10, 20, 30, and 40) in Fig.~\ref{fig:ktinfluence}. The results show that visually meaningful segmentation can be obtained by carefully tuning of the $k_T$. When the $k_T$ is increased, our AF-graph enforces the global structure over the superpixels and masters the meaning of regions. Moreover, our AF-graph preserves more local information in superpixels.

\subsection{Comparison with the state-of-the-art methods}
To verify our framework, we report the quantitative results in comparison with the state-of-the-art methods in Tables~\ref{tab:BSD300}$\sim$\ref{tab:MSRC}. Specially, we highlight in bold the best result for each qualitative metric. The above compared methods include: FH~\cite{Felzenszwalb2004Efficient}, MS~\cite{Comaniciu2002}, Ncut~\cite{JShi200}, MNcut~\cite{1467569}, CCP~\cite{7410546}, Context-sensitive~\cite{4815272}, Corr-Cluster~\cite{Kim2013Task}, SuperParsing~\cite{TIGHE2010Sup}, HIS-FEM~\cite{YIN2017245}, Sobel-AMR-SC~\cite{8733206}, Heuristic better and random better (H\_+R\_Better)~\cite{LiTLL18}, TPG~\cite{6165308}, HO-CC~\cite{6727483}, SAS~\cite{6247750}, $\ell$$_0$-graph~\cite{6738828}, GL-graph~\cite{7024152}, FNCut~\cite{6341755}, Link\_MS+RAG+GLA~\cite{7484679}, SFFCM~\cite{Lei2018fuzzy}, gPb-owt-ucm~\cite{Pablo2011Contour}, RIS+HL~\cite{Wu2014Reverse}, MMGR-AFCF~\cite{8770118}, and AASP-graph~\cite{zhang2019aaspgraph}. For fair comparison, the quantitative results are collected from their evaluations reported in publications.

\begin{table}[!t]
	\renewcommand{\arraystretch}{1.2}
	\caption{Quantitative results of the proposed AF-graph with the state-of-the-art approaches on BSD300 dataset.}
	\centering
	\label{tab:BSD300}
	\begin{tabular}{lcccc}
		\toprule
		Methods                 & PRI $\uparrow$  & VoI $\downarrow$ & GCE $\downarrow$ & BDE $\downarrow$     \\
		\midrule
		Ncut~\cite{JShi200}                   & 0.72 & 2.91 & 0.22 & 17.15  \\
		FCM~\cite{Lei2018fuzzy}               & 0.74 & 2.87 & 0.41 & 13.78  \\
		MNCut~\cite{1467569}                  & 0.76 & 2.47 & 0.19 & 15.10  \\
		SuperParsing~\cite{TIGHE2010Sup}      & 0.76 & 2.04 & 0.28 & 15.05  \\
		HIS-FEM~\cite{YIN2017245}             & 0.78 & 2.31 & 0.22 & 10.66  \\
		SFFCM~\cite{Lei2018fuzzy}             & 0.78 & 2.02 & 0.26 & 12.90  \\
		Context-sensitive~\cite{4815272}      & 0.79 & 3.92 & 0.42 & \textbf{9.90}   \\
		CCP~\cite{7410546}                    & 0.80 & 2.47 &
		\textbf{0.13} & 11.29  \\
		H\_+R\_Better~\cite{LiTLL18}          & 0.81 & 1.83 & 0.21 & 12.16  \\
		Corr-Cluster~\cite{Kim2013Task}       & 0.81 & 1.83 & --   & 11.19  \\
		RIS+HL~\cite{Wu2014Reverse}           & 0.81 & 1.82 & 0.18 & 13.07  \\
		HO-CC~\cite{6727483}                  & 0.81 & 1.74 & --   & 10.38  \\
		TPG~\cite{6165308}                    & 0.82 & 1.77 & --   & --     \\
		SAS~\cite{6247750}                    & 0.83 & 1.65 & 0.18 & 11.29  \\
		$\ell$$_0$-graph~\cite{6738828}       & 0.84 & 1.99 & 0.23 & 11.19  \\
		GL-graph~\cite{7024152}               & 0.84 & 1.80 & 0.19 & 10.66  \\
		AASP-graph~\cite{zhang2019aaspgraph}  & 0.84 & 1.65 & 0.17 & 14.64  \\
		CCP-LAS~\cite{7410546}                & 0.84 & \textbf{1.59} & 0.16 & 10.46 \\
		AF-graph (Linear)                     & \textbf{0.85} & 1.63 & 0.19 & 13.95 \\
		AF-graph (Gaussian, $t=100$)          & \textbf{0.85} & 1.63 & 0.18 & 13.95 \\
		\bottomrule
	\end{tabular}
\end{table}

\begin{table}[!t]
	\centering
	\renewcommand{\arraystretch}{1.2}
	\caption{Quantitative results of the proposed AF-graph with the state-of-the-art approaches on BSD500 dataset.}
	\label{tab:BSD500}
	\begin{tabular}{lcccc}
		\toprule
		Methods   & $\textrm{PRI}\uparrow$    & $\textrm{VoI}\downarrow$   & $\textrm{GCE}\downarrow$   & $\textrm{BDE}\downarrow$  \\
		\midrule
		FCM~\cite{Lei2018fuzzy}               & 0.74 & 2.88 & 0.40 & 13.48  \\
		MNCut~\cite{1467569}                  & 0.76 & 2.33 & --   & --     \\
		MMGR-AFCF~\cite{8770118}              & 0.76 & 2.05 & 0.22 & 12.95  \\
		SFFCM~\cite{Lei2018fuzzy}             & 0.78 & 2.06 & 0.26 & 12.80  \\
		FH~\cite{Felzenszwalb2004Efficient}   & 0.79 & 2.16 & --   & --     \\
		MS~\cite{Comaniciu2002}               & 0.79 & 1.85 & 0.26 & --     \\
		Link\_MS+RAG+GLA~\cite{7484679}       & 0.81 & 1.98 & --   & --     \\
		FNCut~\cite{6341755}                  & 0.81 & 1.86 & --   & --     \\
		Sobel-AMR-SC~\cite{8733206}           & 0.82 & 1.77 & --   & --     \\
		HO-CC~\cite{6727483}                  & 0.83 & 1.79 & --   & \textbf{9.77}  \\
		SAS~\cite{6247750}                    & 0.83 & 1.70 & 0.18 & 11.97  \\
		gPb-owt-ucm~\cite{Pablo2011Contour}   & 0.83 & 1.69 & --   & 10.00  \\
		$\ell_0$-Graph~\cite{6738828}         & 0.84 & 2.08 & 0.23 & 11.07  \\
		AF-graph (Linear)                     & \textbf{0.84} & \textbf{1.67} & \textbf{0.18} & 13.63 \\
		AF-graph (Gaussian, $t=100$)          & \textbf{0.84} & 1.68 & 0.19 & 13.91 \\
		\bottomrule
	\end{tabular}
\end{table}

\begin{table}[!t]
	\centering
	\renewcommand{\arraystretch}{1.2}
	\caption{Quantitative results of the proposed AF-graph with the state-of-the-art approaches on MSRC dataset.}
	\label{tab:MSRC}
	\begin{tabular}{lcccc}
		\toprule
		Methods   & $\textrm{PRI}\uparrow$    & $\textrm{VoI}\downarrow$   & $\textrm{GCE}\downarrow$   & $\textrm{BDE}\downarrow$  \\
		\midrule
		gPb-Hoiem~\cite{Pablo2011Contour}      & 0.61  & 2.85  & --     & 13.53  \\
		MNCut~\cite{1467569}                   & 0.63  & 2.77  & --     & 11.94  \\
		SuperParsing~\cite{TIGHE2010Sup}       & 0.71  & 1.40  & --     & --     \\
		SFFCM~\cite{Lei2018fuzzy}              & 0.73  & 1.58  & 0.25   & 12.49  \\
		Corr-Cluster~\cite{Kim2013Task}        & 0.77  & 1.65  & --     & 9.19   \\
		gPb-owt-ucm~\cite{Pablo2011Contour}    & 0.78  & 1.68  & --     & 9.80   \\
		HO-CC~\cite{6727483}                   & 0.78  & 1.59  & --     & \textbf{9.04}   \\
		RIS+HL~\cite{Wu2014Reverse}            & 0.78  & 1.29  & --     & --     \\
		SAS~\cite{6247750}                     & 0.80  & 1.39  & --     & --     \\
		$\ell_0$-graph~\cite{6738828}          & 0.82  & 1.29  & 0.15   & 9.36   \\
		AF-graph (Linear)             & \textbf{0.83} & 1.24 & \textbf{0.14} & 13.33 \\
		AF-graph (Gaussian, $t=100$)  & 0.82 & \textbf{1.23} & \textbf{0.14} & 13.76 \\
		\bottomrule
	\end{tabular}
\end{table}

\begin{figure}[!t]
	\centering
	\subfloat[$k_T$ = 2]{\includegraphics[width=3.1in]{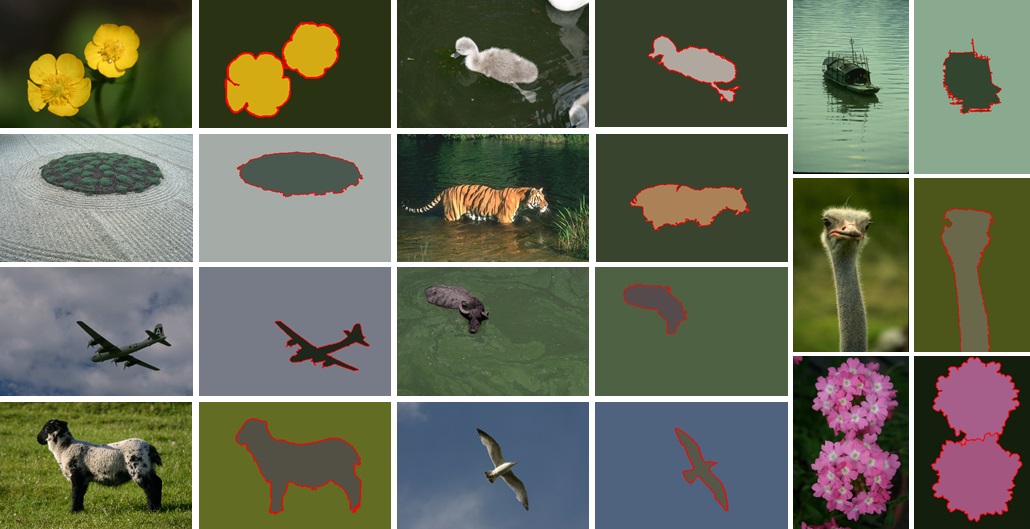}}
	
	\subfloat[$k_T$ = 3]{\includegraphics[width=3.1in]{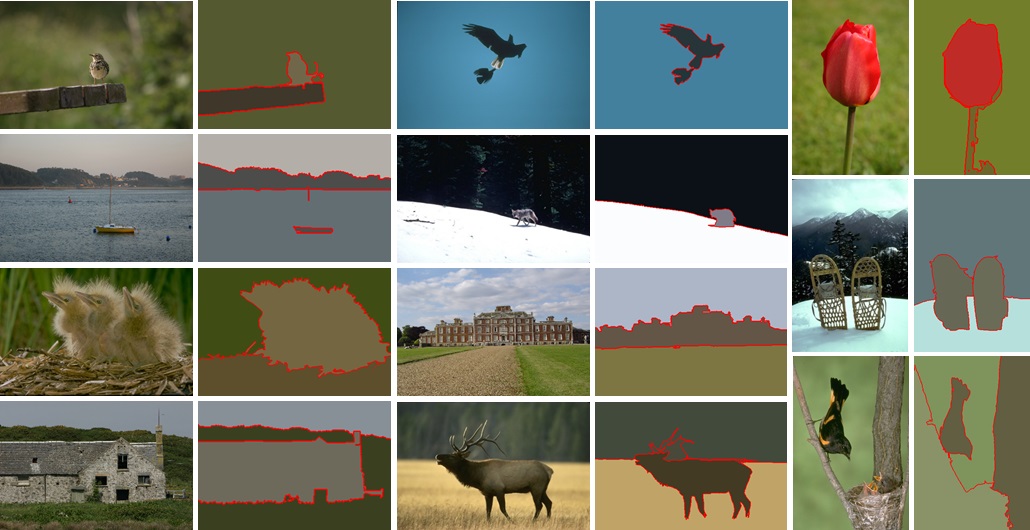}}
	\caption{Visual segmentation results of our AF-graph. All images are segmented into 2 and 3 regions, namely $k_T$ is set to 2 and 3 in Tcut, respectively. Note that salient objects and multiple objects can be segmented accurately.}
	\label{visualres}
\end{figure}

As shown in Table~\ref{tab:BSD300}, our AF-graph ranks the first in PRI and second in VoI on BSD300 dataset. In Table~\ref{tab:BSD500}, our framework achieves the best result in PRI, VoI and GCE on BSD500 dataset. In Table~\ref{tab:MSRC}, our AF-graph ranks the first in PRI, VoI, and GCE on MSRC dataset. To demonstrate the advantages of our AF-graph in practical applications, we present visual segmentation results with $k_T = $ 2 and 3, respectively. From the Fig.~\ref{visualres}, we observe that our framework can be used to segment the salient objects in the following cases: \textbf{i)} the detected object is tiny, such as the airplane, wolf, buffalo, and trawler; \textbf{ii)} multiple objects are needed to be segmented in the same image, such as flower, eagle, boat, and bird; \textbf{iii)} the color of both background and object are quite similar, such as nestling, ostrich, and house.

\begin{figure}[!t]
	\centering
	\includegraphics[width=3.1in]{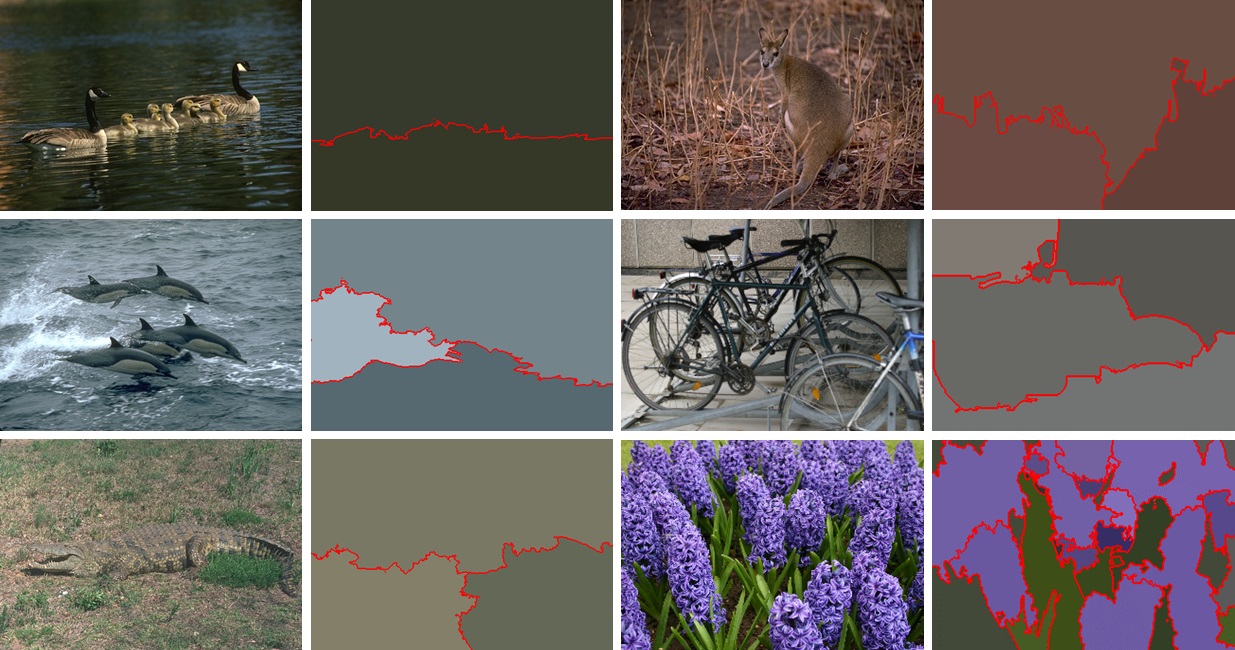}
	\caption{Failure examples by AF-graph. Our AF-graph only uses pixel color information, which fails to capture enough contour and texture cues of segmenting images.}
	\label{fig:falseressult}
\end{figure}

It should be noted that the CCP-LAS (CCP~\cite{7410546} based on layer-affinity by SAS~\cite{6247750}) approach has the most competitive performance due to the integration of the contour and color cues of segmenting images. In contrast, our framework only utilizes the color information. Moreover, the BDE of our framework is unsatisfied. The failure examples by AF-graph are shown in Fig.~\ref{fig:falseressult}. When the detected object is too tiny, and its texture is easily to be confused with background, our framework cannot achieve accurate segmentation. The main reason is that our AF-graph only uses pixel color information, which fails to capture enough contour and texture cues of segmenting images.

\begin{figure*}[!t]
	\centering
	\includegraphics[width=6.9in]{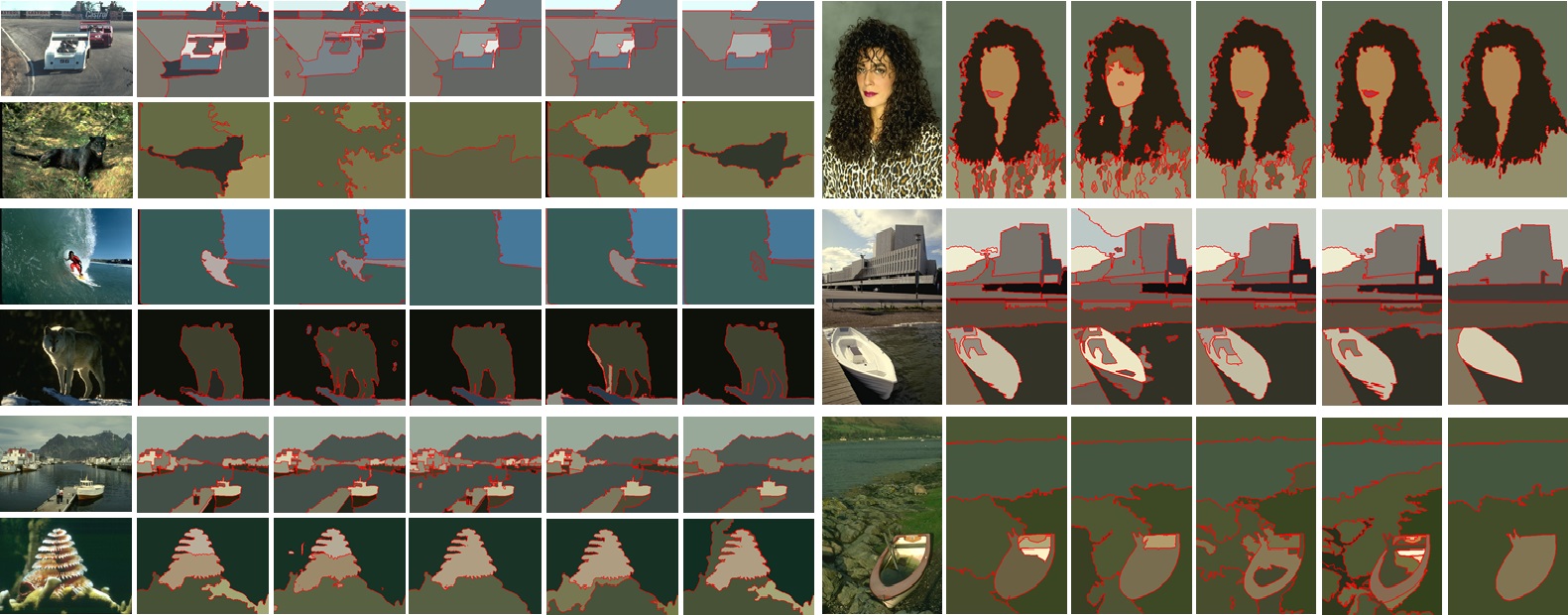}
	\caption{Visual comparison on BSD dataset are obtained by SAS, $\ell$$_0$-graph, GL-graph, AASP-graph, and our AF-graph. Two columns of the comparison results are shown here. From left to right, input images, the results of the SAS, $\ell$$_0$-graph, GL-graph, AASP-graph, and our AF-graph are presented respectively.}
	\label{fig:similarmethod}
\end{figure*}

Especially, our AF-graph follows a similar, but not identical strategy as the SAS, $\ell$$_0$-graph, GL-graph, and AASP-graph. Different from SAS only using adjacent neighborhoods of superpixels and $\ell$$_0$-graph only using $\ell$$_0$ affinity graph of superpixels, our AF-graph can combine different basic graphs. It allows the AF-graph to have a long-range neighborhood topology with a high discriminating power and nonlinearity. The main differences among GL-graph, AASP-graph and our method are the way of graph construction and their fusion principle. In GL-graph, the superpixels are simply classified into three sets according to their areas. In AASP-graph, the superpixels are classified into two sets based on affinity propagation clustering. In our AF-graph, different basic graphs are fused by affinity nodes which are selected by the proposed SSC-SP. Moreover, a novel KSC-graph is built upon these affinity nodes to explore the nonlinear relationships, and then the adjacency-graph of all superpixels is used to update the KSC-graph.

Moreover, various results of the SAS, $\ell$$_0$-graph, GL-graph, AASP-graph, and our AF-graph are shown in Fig.~\ref{fig:similarmethod}, respectively. It shows that our framework achieves a desirable result with less tuning for $k_T$ in Tcut (\emph{e.g.} for surfers, $k_T=3$). The main reason is that our AF-graph selects affinity nodes in an exact way. In particular, our AF-graph achieves the correct and accurate segmentation even in the difficult cases compared with the other similar methods. These cases are: \textbf{i)} the detected object is highly textured, and the background may be highly unstructured (\emph{e.g.} curler, coral, and panther); \textbf{ii)} objects of the same type appear in a large, fractured area of the image (\emph{e.g.} racecars and boat).

\subsection{Time complexity analysis}
\label{sec:time_anal}
Our framework includes the steps of feature representation, graph construction, and graph fusion and partition. Time complexities of SP, KSC-graph construction, Tcut for graph partition are analyzed in Section~\ref{sec:graphconstruction} and Section~\ref{sec:graphpartition} respectively. For each phase, it costs 5.11 seconds to generate superpixels and extract features, 1.65 seconds for affinity nodes selection, 1.68 seconds to build fusion graph, and only 0.82 seconds for graph partition. Our AF-graph takes totally 9.26 seconds to segment an image with the size of 481$\times$321 pixels from BSD on average, which is slower than SAS with 7.44 seconds. In contrast, AASP-graph takes more than 15 seconds in which the global nodes selection and $\ell_0$-graph construction cost much more time than our AF-graph. Moreover, $\ell_0$-graph, MNcut, CCP-LAS, GL-graph, and Ncut usually take more than 20, 30, 40, 100, and 150 seconds, respectively. The main reason is that extracting various features cost too much computational time. All experiments are conducted under the PC condition of 3.40GHz of Intel Xeon E5-2643 v4 processor, 64G RAM, and Matlab 2018a.

\section{Conclusion and future works}
In this paper, our AF-graph combines adjacency-graphs and KSC-graphs by affinity nodes of multi-scale superpixels to obtain a better segmentation result. These affinity nodes are selected by our proposed SSC-SP and further used to construct a KSC-graph. The proposed KSC-graph is then updated by an adjacency-graph of all superpixels at each scale. Experimental results show the good performance and high efficiency of the proposed AF-graph. We also compare our framework with the state-of-the-art approaches, and our AF-graph achieves competitive results on BSD300, BSD500 and MSRC datasets. In the future, we will explore the combination of graph and deep unsupervised learning to improve image segmentation performance.

\ifCLASSOPTIONcaptionsoff
  \newpage
\fi



\bibliographystyle{IEEEtran}
\bibliography{References}


\vfill


\end{document}